\documentclass[journal]{IEEEtran}
\usepackage{graphicx}
\usepackage{comment}
\usepackage[cmex10]{amsmath}
\usepackage{amssymb}
\usepackage{color}

\usepackage{subfigure}

\usepackage{algorithmic}
\usepackage{array}

\begin{document}

\title{Enabling Trustworthy Federated Learning in Industrial IoT: Bridging the Gap Between Interpretability and Robustness}

\author{Senthil Kumar Jagatheesaperumal, Mohamed Rahouti,~\IEEEmembership{Member,~IEEE}, Ali Alfatemi, Nasir Ghani,~\IEEEmembership{Senior Member,~IEEE}, Vu Khanh Quy, and Abdellah Chehri,~\IEEEmembership{Senior Member,~IEEE}

\thanks{S.K. Jagatheesaperumal is with Department of Electronics \& Communication Engineering, Mepco Schlenk Engineering College, Sivakasi, Tamil Nadu, India e-mail: senthilkumarj@mepcoeng.ac.in}
\thanks{M. Rahouti and A. Alfatemi are with the Department of Computer \& Information Science, Fordham University, Bronx, NY USA, 30332 USA e-mail: \{mrahouti; aalfatemi\}@fordham.edu}
\thanks{N. Ghani is with the Department of Electrical Engineering, University of South Florida, Tampa, FL USA (email: nghani@usf.edu)}
\thanks{V. K. Quy is with the Faculty of Information Technology, Hung Yen University of Technology and Education Hung Yen, Vietnam. (e-mail: quyvk@utehy.edu.vn)}

\thanks{A. Chehri is with the Royal Military College of Canada (RMC), Kingston, Ontario, Canada. (e-mail: chehri@rmc.ca)}
}

\markboth{IEEE Internet of Things Magazine}%
{Shell \MakeLowercase{\textit{et al.}}: Bare Demo of IEEEtran.cls for IEEE Journals}

\maketitle

\begin{abstract}
Federated Learning (FL) represents a paradigm shift in machine learning, allowing collaborative model training while keeping data localized. This approach is particularly pertinent in the Industrial Internet of Things (IIoT) context, where data privacy, security, and efficient utilization of distributed resources are paramount. The essence of FL in IIoT lies in its ability to learn from diverse, distributed data sources without requiring central data storage, thus enhancing privacy and reducing communication overheads. However, despite its potential, several challenges impede the widespread adoption of FL in IIoT, notably in ensuring interpretability and robustness. This article focuses on enabling trustworthy FL in IIoT by bridging the gap between interpretability and robustness, which is crucial for enhancing trust, improving decision-making, and ensuring compliance with regulations. Moreover, the design strategies summarized in this article ensure that FL systems in IIoT are transparent and reliable, vital in industrial settings where decisions have significant safety and economic impacts. The case studies in the IIoT environment driven by trustworthy FL models are provided, wherein the practical insights of trustworthy communications between IIoT systems and their end users are highlighted.
\end{abstract}

\begin{IEEEkeywords}
Federated Learning; Industrial IoT; Explainability; Interpretability; Robustness; Environmental Constraints.
\end{IEEEkeywords}

\IEEEpeerreviewmaketitle

\section{Introduction}
\IEEEPARstart{F}{ederated} Learning (FL) has emerged as one of the promising platforms in collaboration with Artificial Intelligence (AI), especially when applied to Industrial Internet of Things (IIoT) networks \cite{vehabovic2023ransomware}. With the core elements of FL, it can be applied to intelligent IIoT systems by facilitating AI training at the network edge. Incorporating FL into IIoT networks makes it possible to address scalability issues while also addressing concerns related to user privacy and the confidentiality of industrial data \cite {nguyen2021federated}.

The integration of FL across diverse industries such as smart cities, healthcare, and transportation is revolutionizing data-sharing methods, attack detection, and privacy preservation. These industries, closely linked with various Internet of Things (IoT) services, benefit from FL's ability to manage data efficiently and securely, as highlighted in Nguyen's study \cite{nguyen2021federatedcomst}. Particularly in IIoT network scenarios, which demand high throughput, low latency, and precise anomaly detection, FL's role becomes even more crucial. By incorporating deep reinforcement learning within an FL framework, FL-based anomaly detection techniques significantly reduce privacy risks, a concept validated in Wang's research \cite{wang2021toward}.

However, the challenge lies in integrating FL, IIoT, and trustworthiness on a single platform, especially within the IIoT context, which is a relatively unexplored area. The challenge is especially due to the tiny footprint of the IIoT devices to withhold the FL model, and subsequently ensure security and privacy for the devices and model. Addressing this, the chameleon hash method with a configurable trapdoor, as proposed in Wei's study \cite{wei2022redactable}, tackles vulnerabilities by employing FL for privacy-preserving data analysis in IIoT. This method has been effectively implemented in a redactable medical blockchain, demonstrating enhanced accuracy and efficiency. Additionally, Li's research introduces a multi-tentacle federated learning (MTFL) architecture for the software-defined IIoT framework, comprising a stochastic tentacle data exchanging (STDE) protocol and an efficient poisoning attack detection algorithm \cite{li2022multitentacle}. These innovations collectively mark significant advancements in the field.

Furthermore, federated cybersecurity, a decentralized approach through FL, plays a pivotal role in identifying and mitigating security threats in the IIoT, as discussed in Ghimire's study \cite{ghimire2022recent}. FederatedTrust, for instance, ensures reliable FL models by employing thirty metrics and six pillars to evaluate trustworthiness in IoT and edge computing, as noted by Sanchez \cite{sanchez2024federatedtrust}. This underscores its practical application in IoT security, particularly in smart industries. Despite these developments, the existing literature on trustworthy FL, especially in IIoT, remains limited, lacking a forward-looking vision for the potential, enabling technologies, challenges, and various design aspects of employing FL in IIoT networks.

The research questions tackled in this paper are outlined as follows:
\begin{itemize}
    \item How to achieve a balance between interpretability and robustness in FL systems within the IIoT, ensuring both transparency for trust and compliance, and resilience against diverse industrial challenges.
    \item What design strategies are effective for creating FL systems in IIoT that are both transparent and reliable, considering the unique industrial requirements where decisions have significant safety and economic impacts?
    \item How can FL be adapted to meet the specific demands and characteristics of IIoT environments, including the aspects of data privacy, security, and efficient utilization of distributed resources?
\end{itemize}
These questions aim to address the complexities of integrating FL into IIoT with a focus on making these systems trustworthy and suitable for industrial applications. To address these questions, we present the following contributions:
\begin{itemize}
    \item We shed light on how to unlock the potential of responsible and robust FL to transform IIoT networks into a self-sustaining architecture.
    \item For the first time in the literature, we provide a holistic overview of the design methodologies and challenging environments of IIoT. Also, we envision the idea of integrating hybrid trustworthy FL design approaches for IIoT networks.
    \item We sketch a road map with four different case studies along with the investigation for the successful realization of trustworthy FL in IIoT-empowered networks.
\end{itemize}

The rest of this paper is organized as follows. The interpretability and robustness enhancements in FL are covered in Section~\ref{sec:Resp}. Section~\ref{sec:Design} presents the design methodologies of FL through deterministic, probabilistic, and adaptive approaches. This is followed by IIoT case studies on manufacturing, energy, supply chain, and environment monitoring in Section~\ref{sec:Case}. Section~\ref{sec:Conc} concludes the paper and highlights future directions.

\section{Responsible and Robust Federated Learning} \label{sec:Resp}

FL has emerged as a crucial technology, enabling decentralized machine learning while addressing key issues like communication efficiency, data privacy, and model accuracy. This is particularly relevant in IIoT, where devices often operate in resource-constrained environments, and the integrity and confidentiality of data are paramount. Recent advancements in FL for IIoT focus on optimizing communication efficiency, data privacy, and model accuracy \cite{boobalan2022fusion}. 
\subsection{FL Enhancement Techniques in IIoT Settings}
\begin{enumerate}
    \item Model compression: This technique reduces the size of the FL models, making them more suitable for transmission over networks with limited bandwidth, which is a common constraint in IIoT environments.
    \item Differential privacy: This approach adds noise to the data or model updates, thus ensuring that individual data points cannot be reverse-engineered, an essential consideration for maintaining data privacy in IIoT.
    \item Secure multi-Party computation: This method enables multiple parties to collaboratively compute a function over their inputs while keeping those inputs private, which is vital in collaborative IIoT settings.
    \item Edge computing paradigms: By processing data close to where it's generated, edge computing reduces latency and bandwidth use, which is crucial for real-time applications in IIoT.
\end{enumerate}

Furthermore, to address the dual challenges of interpretability and robustness in FL, several strategies have been employed:
\subsection{Interpretability Enhancements}
\begin{itemize}
    \item Layer-wise relevance propagation (LRP): It is one of the powerful XAI-based ML approaches that hold the capability to scale towards even the most complex DNNs. In IIoT, where decision-making processes must often be transparent and justifiable, LRP can be particularly valuable. It allows for a detailed understanding of how input data influences the model's output, which is essential for trust in automated systems.
    \item Attention mechanisms: These mechanisms can be specifically tuned to the data types most commonly encountered in IIoT, like time-series sensor data, highlighting the critical aspects for predictions or classifications.
\end{itemize}
\subsection{Robustness Enhancements}
To strengthen the model's robustness against possible attacks, ``adversarial training" entails supplementing training data with adversarial instances that are designed to trick the model. Adversarial training strengthens FL models against malevolent actors trying to impact model integrity in the setting of IIoT, where security risks are common. On the other hand, the term ``byzantine-resistant aggregation" describes methods used in the FL process to reduce the impact of attacks. Byzantine-resistant aggregation guarantees that malicious nodes do not disrupt the collaborative learning process in IIoT settings, where data integrity is crucial. This improves the dependability and credibility of FL models used in IIoT contexts. These methods are essential for preserving FL models' integrity in the face of possible hostile impacts and protecting IIoT systems' efficacy and security.

\begin{itemize}
    \item Adversarial training: In IIoT, where systems can be targets for cyber-attacks, training models with adversarial examples can significantly improve their resilience, ensuring more reliable performance in hostile environments.
    \item Byzantine-resistant aggregation algorithms: Given the distributed nature of IIoT and the potential for compromised nodes, these algorithms are crucial for maintaining the integrity of the FL process, ensuring that malicious or erroneous updates do not skew the model.
\end{itemize}

The advancements in FL for IIoT aim to improve performance and prioritize the systems' robustness and transparency. These qualities are crucial for FL's acceptance and effective operation in industrial environments. The responsible and robust approach to FL in the IIoT sets the stage for developing more secure, efficient, and reliable industrial automation and monitoring systems.

Enhancements in FL for IIoT address particular requirements such as managing heterogeneous data, maintaining privacy, and supporting devices with limited resources. Furthermore, adversarial resilience against cyber-attacks should be given priority by FL approaches, as should the facilitation of real-time adaptation for prompt decision-making. By guaranteeing strong model training, data privacy, and robustness against adversarial assaults in dynamic IIoT contexts, these improvements set apart IIoT-focused FL from other areas.

\begin{figure*}[ht]
  \centering
  \includegraphics[width=0.65\textwidth]{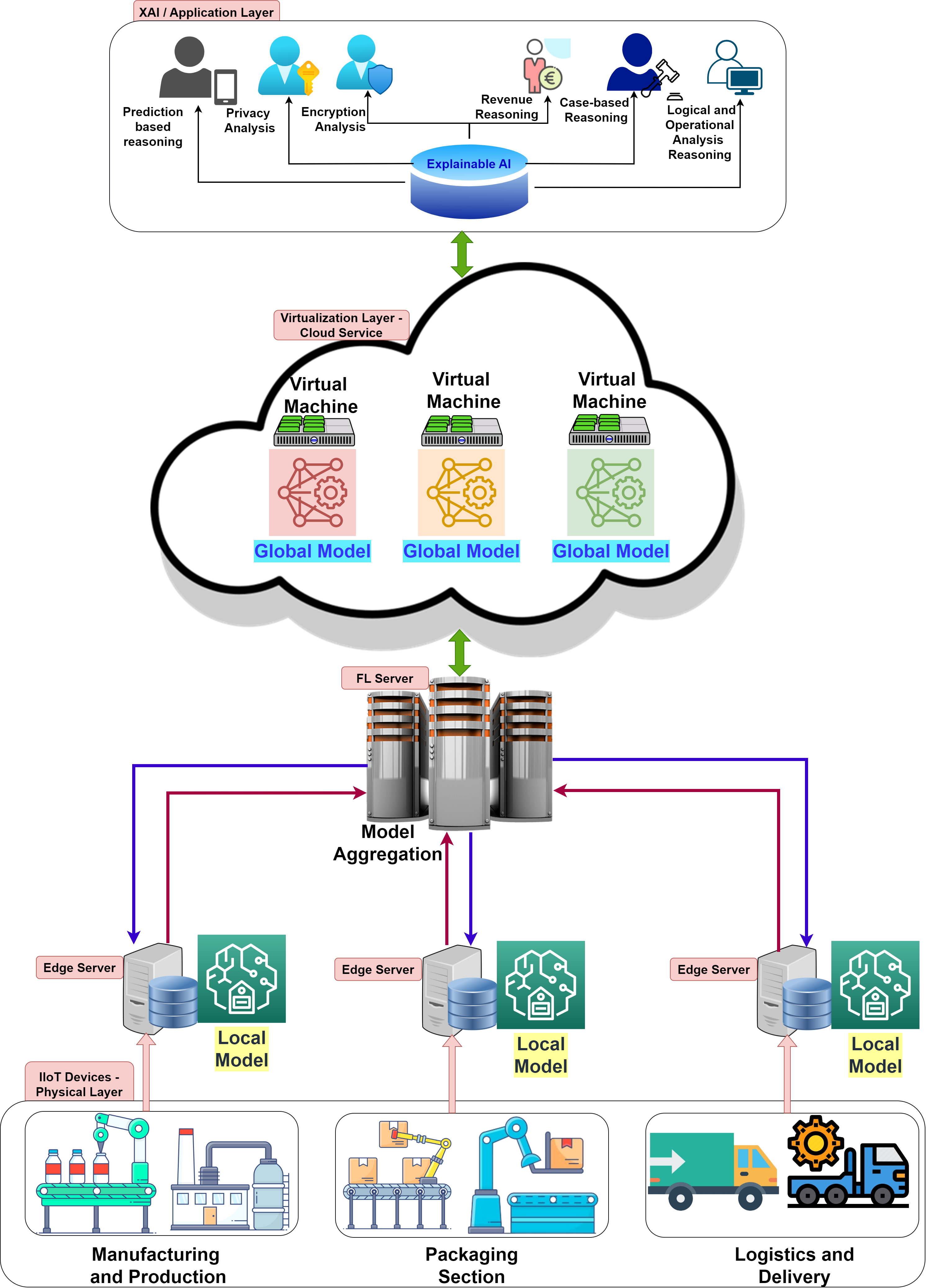}   
  \caption{The Federated Learning Architecture in an Industrial Environment with IIoT-enabled devices and XAI frameworks.}
  \label{fig:FLIIoT}
\end{figure*}

\section{Design Methodologies for Challenging Environments}
\label{sec:Design}

\subsection{Guidelines for Equipping FL Architectures}
Consideration of different emerging technologies in association with FL architectures to handle issues in particularly challenging contexts environments, demands the following general principles and guidelines for improving FL architectures in such settings:

When equipping FL architectures, certain design considerations largely depend on optimizing for IIoT devices with limited resources and managing data heterogeneity using adaptive learning rates and customized preprocessing techniques. Incorporating adversarial robustness techniques like adversarial training and anomaly detection, integrating robust privacy-preserving mechanisms like FL with encryption or differential privacy, and supporting real-time adaptation through online learning and adaptive FL algorithms are also necessary. Furthermore, the design demands lightweight model architectures and edge computing solutions. FL architectures can successfully meet the special requirements of IIoT environments by addressing these issues and design considerations. This allows for collaborative model training while protecting data privacy, guaranteeing model robustness, and promoting prompt decision-making based on dynamic IIoT data streams.

\subsubsection{Robust Communication and Processing}
Implement resilient and decentralized communication protocols with an asynchronous approach, utilizing edge computing for local data processing to cope with unreliable connectivity. Additionally, the incorporation of redundancy and fault-tolerant methods as a proactive measure to effectively manage device failures in challenging scenarios is recommended.

\subsubsection{Resource-Efficient and Adaptive Learning}
Develop FL algorithms that prioritize real-time processing and energy efficiency, considering the constraints of demanding situations. Additionally, it is required to design adaptable models capable of responding to dynamic environments, incorporating reinforcement learning techniques for continuous improvements.

\subsubsection{Security and Privacy Preservation}
Enhance data protection by integrating privacy-preserving technologies such as homomorphic encryption and employing secure communication protocols. Concurrently, optimize data collection by leveraging sensor technologies in collaboration with FL frameworks designed for robustness.

\subsubsection{Continuous Monitoring and Collaboration}
Establish continuous monitoring systems for FL model performance in challenging environments. Subsequently, it is recommended to foster integration among various sensor technologies, implement feedback loops, and evaluate FL structures based on real-world performance data from actual applications.

\subsection{Deterministic Approaches}

Designing robust and efficient systems for challenging environments, particularly in industrial settings, is a complex task that requires innovative and resilient methodologies. This is particularly true when implementing FL in the context of the Industrial IoT and manufacturing processes.

The reliability and performance of FL are put to the test in high-temperature IIoT environments, such as steel manufacturing or chemical processing plants. According to Gaddam et al.~\cite{gaddam2022detecting}, high temperatures can impact hardware performance and data transmission, which can disrupt learning processes. To address these challenges, it is crucial for design methodologies to prioritize the development of hardware and software that can effectively withstand extreme temperatures. This involves the utilization of heat-resistant materials in the construction of devices, as well as the implementation of advanced cooling technologies. On the software side, optimizing algorithms for efficient data processing is crucial to minimize the duration the hardware is exposed to high temperatures. Furthermore, by implementing robust data synchronization methods, it is possible to ensure consistent performance of FL even when there are thermal fluctuations.

In FL for IIoT, decentralization improves dependability by lowering reliance on single points of failure, encouraging redundancy, and lowering data loss risks. Decentralized techniques enhance system resilience, privacy, and security by decreasing vulnerability to centralized weaknesses and aggregating insights locally, notwithstanding the constraints associated with gathering data from remote sources.

In environments where corrosive substances are prevalent, such as in certain types of chemical manufacturing, traditional computing, and IoT devices can degrade quickly, hindering the effectiveness of FL \cite{cenci2022eco}. To address this, design methodologies must include the use of corrosion-resistant materials in device fabrication. Coatings and seals that can withstand harsh chemicals are essential for protecting the devices' internal and external components. From a software perspective, strategies such as redundant data paths and error-checking mechanisms can help maintain data integrity in these harsh conditions. Furthermore, predictive maintenance algorithms integrated into the FL system can preemptively identify and address potential issues caused by corrosive elements, thereby reducing downtime and maintaining steady performance.

In both high-temperature and corrosive environments, the key is to develop FL systems that are resilient to physical challenges and capable of adapting to environmental changes. This involves a continuous cycle of monitoring, learning, and adapting, where the FL system evolves to maintain optimal performance. Such deterministic approaches in the design of FL systems are crucial for their successful deployment and operation in challenging industrial environments.

\subsection{Probabilistic Strategies}
In challenging IIoT environments, which may include factors like varying network conditions, fluctuating data quality, and unpredictable hardware performance, probabilistic models can enhance the resilience of FL systems. These models account for uncertainties and variabilities in the data and operational conditions~\cite{balasubramanian2023fed}. By incorporating probabilistic reasoning, FL systems can make more informed decisions, even in the face of incomplete or noisy data. Techniques like Bayesian inference can be used to update models dynamically as new data becomes available, providing a robust framework for learning under uncertainty.

To further adapt FL to challenging environments, techniques that focus on environmental adaptability are essential. This includes developing adaptive algorithms that can adjust learning parameters in response to changes in the environment. For instance, reinforcement learning strategies can be employed, where the FL system learns optimal actions through trial and error, adapting to the environment's dynamics \cite{khan2021federated}. Furthermore, integrating reliable outlier detection and handling mechanisms guarantees that the FL models are not excessively impacted by abnormal data, which is frequently encountered in unstable environments.

Overall, probabilistic strategies in FL offer a promising avenue for designing systems resilient to the uncertainties and variabilities of challenging IIoT environments. By leveraging these techniques, FL systems can maintain performance and reliability, even under adverse conditions.

\subsection{Adaptive Hybrid Approaches}
The dynamic design fusion in FL involves combining the predictability of deterministic models with the flexibility of probabilistic approaches. Deterministic models provide a solid foundation, offering predictable outcomes under specific conditions, which is vital for ensuring baseline performance and reliability~\cite{boobalan2022fusion}. On the other hand, probabilistic models excel in handling uncertainties and variabilities inherent in IIoT environments, like fluctuating network conditions or variable data quality. By integrating these approaches, FL systems can maintain stable operation while dynamically adapting to changing environmental factors. This fusion allows for robust decision-making, balancing the need for consistency with the ability to respond to new and uncertain information.

To further enhance the robustness of FL in challenging IIoT environments, context-aware methodologies are crucial. These approaches involve designing FL systems that are not only aware of their operational context but can also adapt their learning and decision-making processes based on this context \cite{rentero2022using}. This might include adjusting learning rates, modifying model parameters, or selectively weighting data based on its source and quality. Context-aware systems are particularly adept at dealing with the heterogeneous and dynamic nature of IIoT environments, where the operating conditions can vary widely and unpredictably. By being responsive to the specific context in which they operate, FL systems can optimize their performance and maintain high levels of accuracy and efficiency, even in the most challenging conditions.

These adaptive hybrid approaches in FL design offer a pathway to creating reliable and adaptable systems, capable of thriving in the diverse and often unpredictable landscapes of the IIoT.

In an IIoT setting employing an adaptive hybrid approach to FL integrated with Explainable Artificial Intelligence (XAI) frameworks, the architecture, shown in Figure \ref{fig:FLIIoT}, dynamically combines the benefits of both centralized and decentralized machine learning models. IIoT devices like sensors and assembly machines process and analyze data locally, training AI models with data specific to their environments. This local processing is key in preserving data privacy and minimizing bandwidth usage. Here, the virtualization layer makes resource management and allocation easier, guaranteeing that IIoT devices are used effectively. Model aggregation is handled by FL aggregation servers, and VMs in the cloud offer scalable computational resources. Integration takes place through APIs, which guarantee safe communication and cooperative model training, boosting FL's credibility for IIoT.

In the adaptive hybrid model, these local devices periodically send model updates to a central server, which aggregates these updates to refine a global model. However, unlike traditional FL, this approach can adaptively choose when to rely more on local models and incorporate more from the global model, depending on factors like data diversity, model performance, and network conditions. This adaptability ensures optimal learning outcomes, balancing the need for customized local models with the benefits of a robust, generalized global model.

The integration of XAI in this adaptive hybrid framework is crucial for maintaining transparency and trust in the AI models. XAI tools provide insights into the decision-making processes of both local and global models, ensuring that AI decisions are interpretable and justifiable, a necessity in critical industrial applications. This approach not only enhances operational efficiency and decision-making in IIoT environments but also ensures that AI models remain understandable and trustworthy to human operators, aligning with the evolving needs and constraints of the industrial sector.

\section{Case Studies and Practical Insights}
\label{sec:Case}
\begin{figure*}[ht]
  \centering
  \includegraphics[width=\textwidth]{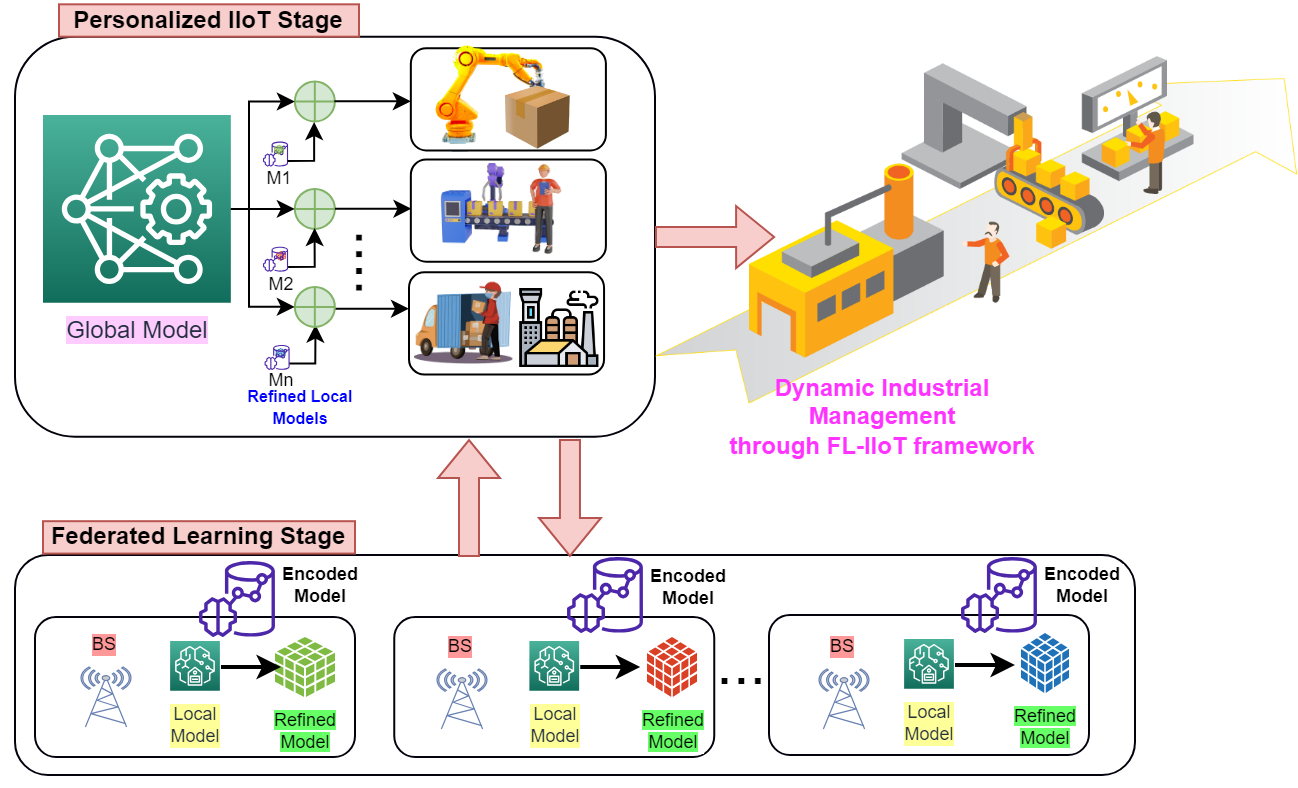}   
  \caption{The Dynamic Industrial Management through Personalized IIoT Framework driven by Federated Learning Models.}
  \label{fig:FLIIoT1}
\end{figure*}

The field of FL in IIoT has seen significant advancements, with real-world applications demonstrating its potential for responsible and robust operations. Below are illustrative case studies and practical insights from various industries. Fig.~\ref{fig:FLIIoT1} shows the dynamic personalized IIoT solutions through secured and trustworthy FL models.

\subsection{Smart Manufacturing}
In a smart manufacturing scenario, FL is utilized to enhance production processes by optimizing them, all while prioritizing data privacy. FL can identify patterns and inefficiencies by analyzing data across multiple manufacturing sites, all while ensuring that sensitive information is not shared. For example, a multinational company can utilize FL to streamline its quality control procedures across various locations, resulting in improved efficiency and decreased waste.

\subsection{Energy Sector}
In the energy sector, particularly in renewable energy, FL is applied to predict maintenance needs and optimize energy distribution. By processing data from various wind farms or solar panels, FL can predict when maintenance is required, thereby reducing downtime and maximizing energy output. This application not only improves operational efficiency but also contributes to the sustainability of energy systems.

\subsection{Supply Chain Management}
FL can significantly enhance supply chain management by providing insights into logistics without compromising the data of individual suppliers. For example, in a complex supply chain, FL can be used to track and predict inventory levels, ensuring timely restocking and reducing the risk of supply shortages.

\subsection{Environmental Monitoring}
In environmental monitoring, FL is used to process data from various sensors deployed in different geographical locations. This data helps in predicting environmental trends and potential hazards, such as pollution levels or the likelihood of natural disasters. Such applications are crucial for proactive environmental protection and disaster management.

These case studies demonstrate the versatility and efficacy of FL in IIoT, highlighting its ability to offer robust solutions while adhering to responsible data-handling practices. As the technology continues to evolve, it is expected to find even more innovative applications across different sectors.

The future of trustworthy FL in IIoT lies in developing holistic frameworks that integrate robustness and interpretability seamlessly. Potential directions include:
\begin{itemize}
    \item XAI integration: More profound integration of XI techniques within FL models to enhance transparency and trust.
    \item Advanced anomaly detection: Leveraging AI to detect and mitigate novel attacks and anomalies in FL networks.
    \item Hybrid models: Combining the strengths of different learning paradigms (supervised, unsupervised, and reinforcement learning) within the FL framework for more robust and interpretable models.
\end{itemize}

\subsection{Practical Insights}

FL provides customized solutions for improving industrial processes in challenging industrial ecosystems. By cooperatively enhancing pressure control among dispersed devices, FL can improve low-pressure processes in the chemical and pharmaceutical sectors. FL is also useful for evaluating equipment health in low-temperature conditions and optimizing temperature control in cryogenic industries, such as the manufacturing of liquefied natural gas. FL helps offshore oil and gas operations improve overall safety in challenging and remote environments and guarantee the quality of subsea equipment. Extreme temperature and pressure changes are addressed by FL in the aerospace manufacturing industry, which also optimizes furnace conditions and forecasts equipment failures to boost productivity. FL preserves data privacy, minimises communication overhead, and allows collaborative model training directly on distributed IIoT devices, providing unique advantages for evaluating equipment health in low-temperature environments. This strategy preserves the confidentiality of sensitive data and guarantees real-time insights, which is essential for preserving equipment dependability under challenging conditions.

Furthermore, FL is also used in mining operations to optimize autonomous vehicles, track the condition of equipment, and improve extraction procedures in difficult terrain. Such applications show the flexibility of FL in meeting certain industry requirements while maintaining data security and privacy.

Enabling trustworthy FL in the context of IIoT presents a unique set of challenges and opportunities, primarily centered around balancing interpretability and robustness. FL, a decentralized approach to machine learning, allows for the training of algorithms across multiple devices or servers while keeping the data localized. This method is particularly beneficial for IIoT, where data security and privacy are paramount, and the data is often generated in distributed environments \cite{ur2021trustfed}.

One key lesson learned is the importance of interpretability in FL models. In the industrial setting, stakeholders need to understand and trust the decisions made by AI models, especially when these decisions can have significant safety and financial implications. Interpretability in FL can be challenging due to the distributed nature of the model training, but it's crucial for ensuring that the models are reliable and their decisions are justifiable. Techniques such as model-agnostic methods, visualization tools, and simplified model architectures have been explored to enhance interpretability without compromising the model's performance.

On the other hand, robustness in FL is essential for maintaining the integrity and performance of the models in the face of diverse and often noisy industrial data. Robustness involves ensuring that the models are resistant to data anomalies, adversarial attacks, and other forms of disruptions that are common in IIoT environments. This requires advanced techniques like anomaly detection, data encryption, and secure aggregation protocols in the FL framework. The research in this area focuses on developing robust FL models that can withstand the challenges specific to industrial settings, such as variable data quality and potential cybersecurity threats.

An XAI model predicting machine failures through FL and IIoT frameworks in a manufacturing plant could provide interpretable features such as vibration levels, temperature fluctuations, and health status to diagnose unexpected issues, aiding in understanding the root causes of equipment failures and enabling proactive maintenance actions. Furthermore, robust XAI models ensure the consistency of product quality by addressing quality variations in materials and environmental conditions. The interpretability feature allows supply chain managers to understand better the factors affecting performance and identify areas for improvement. Additionally, the robustness of the models helps manage supply chain disruptions, demand volatility, and unforeseen market changes. In packaging materials, interpretability allows inspectors to grasp the criteria used by XAI algorithms to detect defects and anomalies. To uphold product quality and brand reputation, the robustness of the model assists in estimating variations in surface textures, defects, and packaging configurations.

Last, the integration of trustworthy FL in IIoT hinges on achieving a balance between interpretability and robustness. While interpretability ensures the transparency and trustworthiness of the models, robustness safeguards them against the diverse challenges in industrial environments. Ongoing research and practical applications are geared towards developing methodologies and tools that enhance both aspects, thereby making FL a viable and reliable approach in the context of IIoT.

\section{Conclusion} \label{sec:Conc}
While FL allows learning from diverse, distributed data sources without central data storage, enhancing privacy and reducing communication needs, its widespread adoption in IIoT is challenged by issues ensuring model interpretability and robustness. Trustworthy FL in the IIoT is a complex yet promising domain. Bridging the gap between interpretability and robustness requires a multifaceted approach involving advanced machine-learning techniques, robust security protocols, and an overarching framework prioritizing transparency and resilience. As technology evolves, the synergy of FL with IIoT will undoubtedly play a pivotal role in realizing the full potential of smart industries.


\ifCLASSOPTIONcaptionsoff
  \newpage
\fi

\begin{IEEEbiographynophoto}{Senthil Kumar Jagatheesaperumal}
received his B.E. degree in Electronics and Communication Engineering from Madurai Kamaraj University, Tamilnadu, India in 2003. He received his post-graduation degree in Communication Systems from Anna University, Chennai, India in 2005. He completed his Ph.D. in Information and Communication Engineering from Anna University, Chennai, India in 2017. He is currently working as an Associate Professor (Senior Grade) in the Department of Electronics and Communication Engineering, Mepco Schlenk Engineering College, Sivakasi, Tamilnadu, India. He received two funded research projects from National Instruments, USA each worth USD 50,000 during the years 2015 and 2016. He also received another funded research project from IITM-RUTAG in 2017 worth Rs.3.97 Lakhs. His area of research includes Robotics, the Internet of Things, Embedded Systems, and Wireless Communication. He is a Life Member of IETE and ISTE.
\end{IEEEbiographynophoto}

\begin{IEEEbiographynophoto}{Mohamed Rahouti}
received the M.S. and Ph.D. degrees from the University of South Florida in Mathematics Dept. and Electrical Engineering Dept., Tampa, FL, USA, in 2016 and 2020, respectively. He is currently an Assistant Professor, Department of Computer and Information Sciences, Fordham University, Bronx, NY, USA. His current research focuses on computer networking, blockchain technology, Internet of Things (IoT), machine learning, and network security with applications to smart cities.
\end{IEEEbiographynophoto}

\begin{IEEEbiographynophoto}{Ali Alfatemi}
is pursuing his Ph.D. in Computer science at Fordham University, USA. He has extensive industry experience as a Software Engineer. His research interests encompass Artificial Intelligence and Machine Learning, Network Intrusion Detection, and Bio-informatics.  
\end{IEEEbiographynophoto}

\begin{IEEEbiographynophoto}{Nasir Ghani}
is a Professor of Electrical Engineering and Program Director of the College of Engineering MS in Cybersecurity. He is also Academic Research Director for Cyber Florida, a state-funded cybersecurity center focusing on research, education, and outreach. Earlier, he was Associate Chair of the ECE Department at the University of New Mexico (UNM). He has also held technical research and development positions at several large corporations (including Nokia, IBM, and Motorola) and some startups. His research focus areas include cybersecurity, cyberinfrastructure design, disaster recovery, and online education. His research has been supported by the NSF, DoD, DoE, Qatar Foundation, and several state and industry partners. He also received the NSF CAREER Award in 2005.
\end{IEEEbiographynophoto}

\begin{IEEEbiographynophoto}{Vu Khanh Quy}
was born in Hai Duong, Vietnam, in 1982. He received the M.Sc. and Ph.D. degrees from the Posts and Telecommunications Institute of Technology, Hanoi, Vietnam, in 2012 and 2021, respectively. He is currently a Lecturer with the Hung Yen University of Technology and Education (UTEHY), Hung Yen, Vietnam. His research interests include wireless communications, mobile computing, smart IoT systems, next-generation networks, Internet of Things, and next-communication networks
\end{IEEEbiographynophoto}

\begin{IEEEbiographynophoto}{Abdellah Chehri}
(Senior Member, IEEE) is an Associate Professor at the Royal Military College of Canada (RMC), Kingston, Canada. Before joining the RMC, he was an associate professor at the University of Quebec (UQAC). He has an affiliate professor at the University of Quebec UQO, UQAC and an adjunct professor at the University of Ottawa. He has served as guest/associate editor for several well-reputed journals. Dr. Chehri is a Senior Member of IEEE, a member of the IEEE Communication Society (ComSoc), IEEE Vehicular Technology Society (VTS), and IEEE Photonics Society, 
\end{IEEEbiographynophoto}

\end{document}